\documentclass[runningheads]{llncs}

\usepackage[T1]{fontenc}
\usepackage{graphicx}
\usepackage{hyperref}
\usepackage{array}
\usepackage{subcaption}

\begin{document}

\title{Emotion Classification In-Context in Spanish}
%
%
\author{Bipul Thapa\inst{1} \and
Gabriel Cofre\inst{2}
}
\authorrunning{B. Thapa and G. Cofre}
%
\institute{Department of Computer and Information Sciences, University of Delaware, Newark, DE, USA \\
\email{bipul@udel.edu}\\ \and
Department of Physics and Astronomy, University of Delaware, \\ Newark, DE, USA\\
\email{gcofre@udel.edu}}
\maketitle

\begin{abstract}
Classifying customer feedback into distinct emotion categories is essential for understanding sentiment and improving customer experience. In this paper, we classify customer feedback in Spanish into three emotion categories—positive, neutral, and negative—using advanced NLP and ML techniques. Traditional methods translate feedback from widely spoken languages to less common ones, resulting in a loss of semantic integrity and contextual nuances inherent to the original language. To address this limitation, we propose a hybrid approach that combines TF-IDF with BERT embeddings, effectively transforming Spanish text into rich numerical representations that preserve the semantic depth of the original language by using a Custom Stacking Ensemble (CSE) approach. To evaluate emotion classification, we utilize a range of models, including Logistic Regression, KNN, Bagging classifier with LGBM, and AdaBoost. The CSE model combines these classifiers as base models and uses a one-vs-all Logistic Regression as the meta-model. Our experimental results demonstrate that CSE significantly outperforms the individual and BERT model, achieving a test accuracy of~93.3\% on the native Spanish dataset—higher than the accuracy obtained from the translated version. These findings underscore the challenges of emotion classification in Spanish and highlight the advantages of combining vectorization techniques like TF-IDF with BERT for improved accuracy. Our results provide valuable insights for businesses seeking to leverage emotion classification to enhance customer feedback analysis and service improvements.

\end{abstract}
\section{Introduction}
With the recent advancements in Artificial Intelligence and Machine Learning (ML), these technologies are at the forefront of driving innovation in developing next-generation intelligent systems. Their impact is especially notable in Natural Language Processing (NLP), where AI and ML techniques are crucial for extracting and interpreting human emotions from text. This enhances the understanding of user emotions and improves personalized user experiences. 

Emotion classification is a crucial component of NLP, playing an essential role in understanding and interpreting effective states from textual data, which is key to sentiment analysis and emotion recognition~\cite{nandwani2021review}. As digital communication continues to grow, especially through social media and customer feedback channels, the need to accurately analyze emotions embedded within text has become increasingly important. Spanish, being one of the widely spoken language systems, presents unique challenges and opportunities for emotion classification due to its linguistic complexities and diverse regional variations~\cite{lin@2022}. Thus, effective emotion classification in Spanish becomes essential for businesses and organizations aiming to tap into Spanish-speaking markets and improve their services based on customer feedback.

However, the current trend involves translating datasets from widely available languages into less commonly represented ones due to the limited availability of native datasets in these languages and subsequently training models on the translated data. This approach often fails to capture the semantic meaning and contextual dependencies natural to the original language, leading to poor accuracy performance. For instance, in the case of \href{https://grehus.com/}{Grehus} dataset, existing libraries (such as 'polarity') demonstrate inadequate performance when trained, achieving an accuracy of around~30\% only, as shown in Table~\ref{tab:classification_report}. Therefore, utilizing the original Spanish dataset is necessary to achieve higher accuracy in emotion classification for the Spanish language. Furthermore, the selection of appropriate models and feature representations is also crucial to ensuring reliable and accurate emotion classification.

\begin{table}
\caption{Classification Report for Sentiment Analysis with Polarity}\label{tab:classification_report}
\centering 
\begin{tabular}{|c|c|c|c|}
\hline
{\bfseries Class} & {\bfseries Precision} & {\bfseries Recall} & {\bfseries F1-score} \\ 
\hline
Positive & 0.55 & 0.02 & 0.03 \\
\hline
Neutral & 0.29 & 0.99 & 0.45 \\
\hline
Negative & 0.33 & 0.00 & 0.01 \\
\hline
\multicolumn{4}{|c|}{\bfseries Overall Accuracy: 29.7\%} \\
\hline
\end{tabular}
\end{table}

The objective of this paper is to classify customer feedback in Spanish by using advanced NLP techniques and ML models to accurately categorize emotions into three distinct groups: positive, neutral, and negative. Our work includes a comprehensive analysis of Spanish customer feedback data. By focusing on native Spanish data rather than translated datasets, we aim to overcome the limitations of translation.

\footnotetext[1]{Available: \href{https://www.grehus.com/}{https://www.grehus.com/}}

Our approach involves a multi-step process to ensure robust emotion classification. We use a dataset containing responses to open-ended questions from the healthcare industry, collected from the GreHus company~\footnotemark[1]. To our knowledge, this dataset has not been previously utilized for emotion classification tasks, which contributes to its uniqueness and novelty in this research context. These open-ended questions aim to capture a wide range of customer sentiments, including motivations, expectations, and suggestions for improvement. The dataset is preprocessed to clean and prepare the text for analysis, ensuring that it is in a suitable format for ML algorithms.

For model training and evaluation, we implement four different machine learning models: Logistic Regression, K-Nearest Neighbors (KNN), Bagging classifier with Light Gradient-Boosting Machine (LGBM), and AdaBoost, along with a custom ensemble model called the Custom Stacking Ensemble~(CSE). The CSE model has Logistic Regression, KNN, Bagging classifier with LGBM, and Adaboost as the base model, with a one-vs-rest Logistic Regression as the meta-model. We train all of these models on the Spanish dataset and evaluate them using metrics such as accuracy, precision, recall, and F1-score to determine their effectiveness in classifying emotions. To expand our analysis and understand the effect of translation on emotion classification, we also use an English-translated version of the dataset, comparing model performance between the native Spanish text and its translated counterpart. This comparison reveals how translation influences the accuracy and effectiveness of emotion classification. To enhance the text representation for both datasets, we combine Term Frequency-Inverse Document Frequency (TF-IDF) and Bidirectional Encoder Representations from Transformers (BERT) embeddings and evaluate the results across the models.

By directly leveraging native Spanish data, we aim to improve the accuracy and reliability of emotion classification, providing businesses with more precise insights into customer sentiment. This, in turn, supports businesses in optimizing their services and enhancing customer satisfaction. Our key contributions are summarized as follows:
\begin{itemize}
    \item We introduce a Custom Stacking Ensemble (CSE) model that integrates TF-IDF and BERT embeddings with stacking ensemble models, enabling effective emotion classification for Spanish customer feedback.
    \item We utilize native Spanish healthcare industry datasets, which have not been previously explored, and provide valuable insights into the impact of native versus translated datasets, highlighting the benefits of using native language data for improved sentiment analysis.
    \item We demonstrate the efficiency of the proposed approach by achieving an accuracy~93.3\% on native Spanish datasets, outperforming traditional baselines and transformer-based models.
    \item We perform a comprehensive comparison of traditional machine learning models, ensemble methods, and fine-tuned transformer models, underscoring the practicality and performance efficiency of the CSE approach.
\end{itemize}

The rest of the paper is organized as follows. In Section~\ref{sec:realated-work}, we provide a review of related work in this field. Section~\ref{sec:method} details the methodology and proposed model, including the machine learning approaches and feature representation techniques used in this research. In Section~\ref{sec:experiments}, we present an extensive experimental evaluation of the models, comparing performance metrics across various approaches. Finally, in Section~\ref{sec:conclusion}, we summarize our findings and discuss potential directions for future research.

\section{Related Work}
\label{sec:realated-work}
Previous work in sentiment analysis has employed various approaches, from lexicon-based methods to machine-learning models. Lexicon-based methods, such as the Sentiment Polarity Library~\cite{loria2018textblob}, assign polarity scores to words and aggregate these scores to determine the overall sentiment of a text. Polarity is the basic and frequently used categorization for sentiment analysis that often specifies whether a comment is positive or negative~\cite{molina2013semantic}. However, these methods often fall short of capturing the nuanced expressions of emotions, particularly in Spanish text.

As shown in Table~\ref{tab:classification_report}, sentiment analysis using the polarity library on our dataset yields significant limitations, especially in detecting positive and negative sentiments, with F1-scores of~0.03 and~0.01, respectively. This underscores the need for more sophisticated approaches that can better capture the intricate expressions of emotions in Spanish.

TF-IDF is a widely adopted feature selection method frequently utilized in diverse applications, including sentiment analysis, fake news classification, and numerous other natural language processing tasks~\cite{bhowmik2022novel}~\cite{alqaryouti2024aspect}~\cite{ahuja2019impact}~\cite{aurpa2022abusive}. While TF-IDF has been effective for many NLP tasks, recent advances in text vectorization have introduced state-of-the-art techniques such as Word2Vec, GloVe, Context2Vec, CoVe, and transformer-based models like BERT, GPT, and RoBERTa, which have significantly improved performance in these tasks~\cite{patil2023survey}~\cite{asudani2023impact}~\cite{mutinda2023sentiment}. However, the integration of traditional methods with modern embeddings has not been fully explored, particularly when applied to Spanish datasets. Our work addresses this gap by employing a hybrid feature representation, combining TF-IDF and BERT.

Additionally, various studies have extensively utilized transformer-based models for sentiment analysis. Some approaches leverage transformer models primarily for generating embeddings, while others employ them directly for classification tasks~\cite{naseem2020transformer}~\cite{farahani2021parsbert}\cite{acheampong2021transformer}~\cite{kokab2022transformer}. In contrast, our work utilizes transformer-generated embeddings as input features and applies a simpler yet effective ensemble model for sentiment classification, striking a balance between performance and computational efficiency.

ML approaches, including traditional models such as Naive Bayes, SVM, as well as deep Learning models like LSTM, CNN, BiLSTM, have shown promise in sentiment analysis~\cite{wongkar2019}~\cite{hameed2020sentiment}~\cite{xu2020deep}~\cite{joshi2022deep}~\cite{olusegun2023text}. However, most studies focus on English datasets, with limited research on Spanish, where translation methods often fail to capture contextual nuances, reducing accuracy. Also, deep learning models are resource-intensive and require large datasets for higher accuracy. Our work bridges these gaps by leveraging native Spanish datasets and advanced NLP techniques with a simple yet efficient ensemble approach to improve emotion classification accuracy. 


\section{Method}
\label{sec:method}

\begin{figure}[h]
\centering
\includegraphics[width=0.7\linewidth]{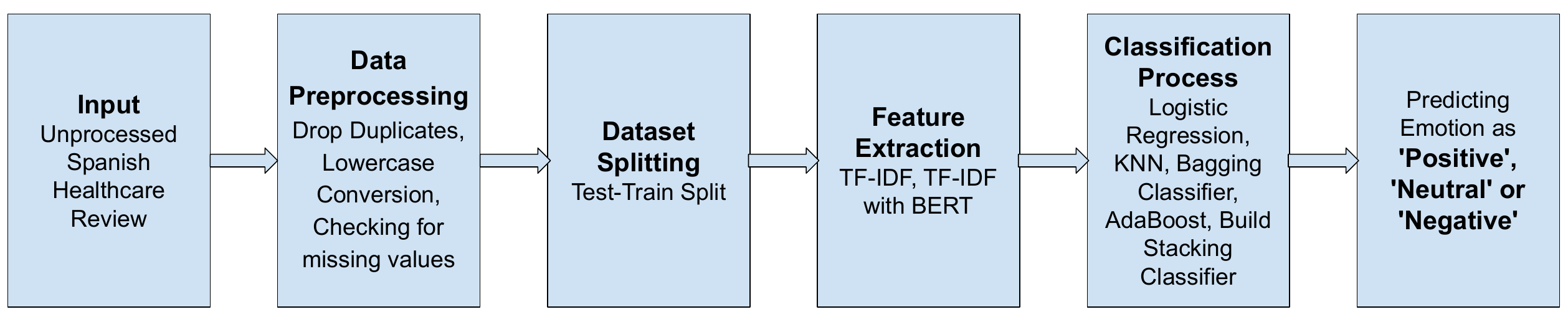}
    \caption{Overview of the method}
    \label{fig:overview}
\end{figure}

In our work, we employ a comprehensive multi-step methodology to accurately classify emotions within Spanish customer feedback as shown in Figure~\ref{fig:overview}. The approach encompasses data preprocessing, vectorization techniques, model training, evaluation, and performance assessment. Each stage is meticulously designed to ensure robustness and effectiveness.

\subsection{Data Preparation}
The initial step involves preparing the data by collecting a dataset containing textual responses from the healthcare industry, specifically from the GreHus company. The data collected are in Spanish. A sample of this dataset is shown in Table~\ref{tab:sentiment_classification}. The dataset captures a wide range of customer sentiments through open-ended questions related to motivations, expectations, and suggestions for improvement.

\begin{table}[htbp]
\caption{Spanish Dataset Sample}
\begin{center}
\begin{tabular}{|c|c|}
\hline
\textbf{Customer Feedback} & \textbf{Emotion} \\ 
\hline
La comodidad y la buena atención & Positive \\
\hline
Facilidades de buen servicio & Positive \\
\hline
Los precios & Neutral \\
\hline
Falta de tiempo & Neutral \\
\hline
Mejor precio & Negative \\
\hline
ME QUEDE ESPERANDO LA LLAMADA & Negative \\
\hline
\end{tabular}
\label{tab:sentiment_classification}
\end{center}
\end{table}

The data preparation phase, as depicted by the first three boxes in Figure~\ref{fig:overview}, involves Input, Data Preprocessing, and Data Splitting. To facilitate data analysis, various preprocessing steps are carried out. These include removing duplicate entries, converting text to lowercase, and handling missing values. These steps help to minimize noise, improve data quality, and ensure uniformity throughout the dataset.

\subsection{Feature Extraction}
Accurate representation of textual data is crucial for effective model performance. We employ
two complementary feature extraction techniques to transform the textual data into numerical representations suitable for ML algorithms, TF-IDF and BERT embeddings.  

TF-IDF captures the importance of words within the documents by calculating the frequency of each word relative to its occurrence across all documents. Technically, this technique transforms textual data into a sparse numerical matrix by computing the term frequency (TF) and inverse document frequency (IDF) for each word. TF-IDF is particularly effective for identifying discriminative words across the dataset, ensuring that commonly used words are assigned lower weights.

BERT embeddings~\cite{devlin2018bert} capture contextual relationships and semantic meaning by encoding the entire sentence. BERT provides rich contextual embeddings for each word in the text. Unlike traditional embeddings, BERT accounts for word position and co-occurrence, enabling a richer representation of emotions in feedback.

We combine the TF-IDF and BERT embeddings into a single feature vector. By combining TF-IDF and BERT vectors, the strengths of both techniques are leveraged~\cite{jin2020multi}~\cite{sun2022text}, enhancing the model's ability to understand and classify emotions. The hybrid approach captures both shallow statistical relationships and deep semantic structures, providing a richer feature space for classification.

\subsection{Classification Process}
Several ML models are implemented and assessed to classify emotions in the customer feedback data. These models include Logistic Regression, a Bagging classifier with an LGBM, KNN, AdaBoost, and a CSE approach.

Logistic Regression is selected as a baseline model due to its simplicity and effectiveness in handling linear decision boundaries. It provides an interpretable and straightforward benchmark for evaluating more complex models on the Spanish dataset.

The Bagging classifier with LGBM is chosen for its efficiency in handling high-dimensional data. By using LGBM with bagging, this model effectively reduces variance, making it suitable for the complex feature interactions present in the Spanish dataset.

KNN is included for its ability to capture non-linear relationships in the data. By considering local information in the feature space, KNN offers a different perspective compared to the gradient boosting and logistic regression models, which is particularly useful for classifying complex patterns in the Spanish dataset.

AdaBoost is selected for its capability to improve accuracy by focusing on difficult-to-classify instances. This model is particularly effective in handling noisy data and class imbalances, making it a valuable addition to evaluating the Spanish dataset.

\begin{figure}[h]
    \centering
    \includegraphics[width=0.7\linewidth]{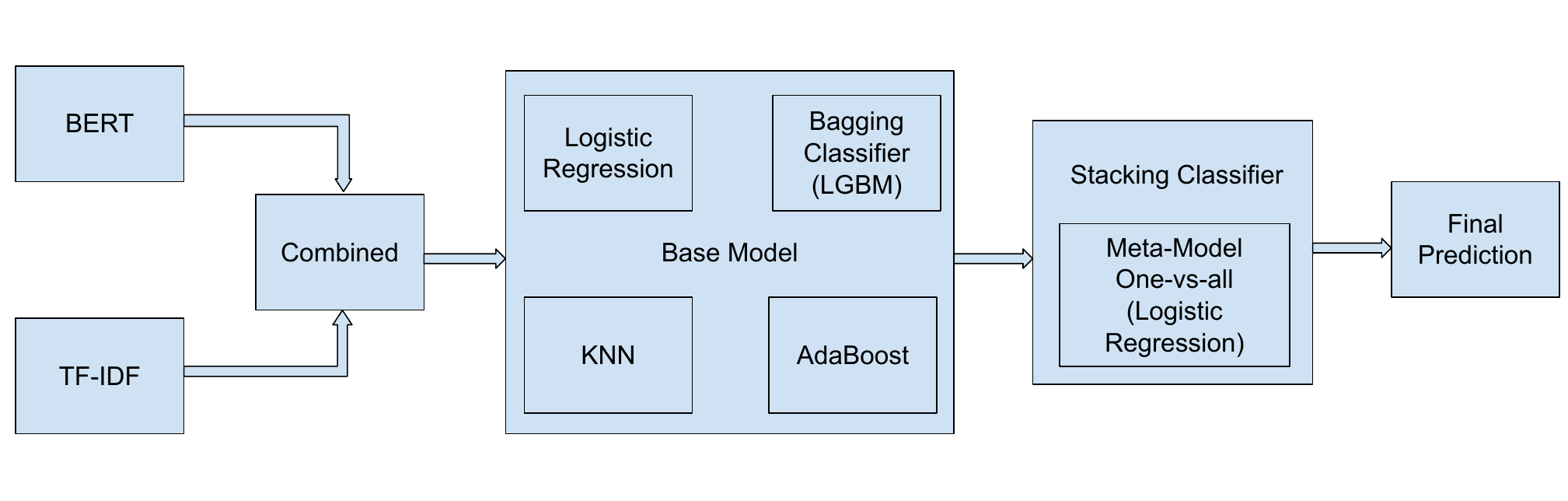}
    \caption{CSE approach}
    \label{fig:cse_approach}
\end{figure}

\subsubsection{Custom Stacking Ensemble (CSE)}
In this work, we propose a CSE approach that integrates multiple base classifiers into a unified predictive model. The base classifiers employed in the ensemble include Logistic Regression, LGBM, KNN, and AdaBoost, each selected for its unique strengths and complementary characteristics as shown in Figure~\ref{fig:cse_approach}. These base models are individually trained on the dataset, and their predictions serve as input features for the meta-classifier.

We perform a grid search to tune the key hyperparameters of all base classifiers and the meta-classifier, including the regularization strength for Logistic Regression, the number of estimators for LGBM and AdaBoost, and the number of neighbors for KNN, as presented in Table~\ref{tab:grid_search_params} for the CSE.

\begin{table}[ht]
\centering
\caption{Hyperparameter Grid Search and Best Parameters for CSE}
\label{tab:grid_search_params}
\begin{tabular}{|l|c|c|}
\hline
\textbf{Parameter}                     & \textbf{Grid Values}               & \textbf{Best Value} \\ \hline
logreg\_\_C                   & \{0.01, 0.1, 10\}                  & 0.1 \\ \hline
lgbm\_\_n\_estimators         & \{50, 100, 200\}                   & 50 \\ \hline
lgbm\_\_learning\_rate        & \{0.01, 0.1, 0.2\}                 & 0.01 \\ \hline
knn\_\_n\_neighbors           & \{3, 5, 7\}                        & 3 \\ \hline
adaboost\_\_n\_estimators     & \{50, 100, 200\}                   & 50 \\ \hline
final\_estimator\_\_estimator\_\_C & \{0.01, 0.1, 1, 10\}             & 0.1 \\ \hline
\end{tabular}
\end{table}

The meta-classifier in the CSE architecture is one-vs-rest Logistic Regression, chosen for its effectiveness in multi-class classification tasks. This strategy decomposes the multi-class problem into several binary classification problems, where a separate logistic regression model is trained to distinguish each class from all others. By simplifying the multi-class problem into multiple binary sub-problems, the one-vs-rest approach enhances the ability of the stacking ensemble to capture complex decision boundaries in high-dimensional feature spaces.

To ensure robustness and mitigate the risk of overfitting, the CSE model is evaluated with training data using 5-fold cross-validation. This procedure divides the dataset into five partitions, with the model being trained on four partitions and validated on the remaining one. This process is repeated five times, with each partition serving as the validation set once. The use of cross-validation ensures that the reported performance metrics are reliable and generalizable across different subsets of the data.

By combining the predictive power of multiple base classifiers and optimizing their integration through a one-vs-rest Logistic Regression meta-model, the proposed CSE approach demonstrates improved performance and generalization over individual models.

\subsection{Baseline Comparison}
We separately use the Logistic Regression, Bagging Classifier with LGBM, KNN, and AdaBoost for the sentiment analysis and compare with our CSE approach. Alongside these, we also fine-tune the pre-trained "nlptown/bert-base-multilingual-uncased-sentiment"~\footnotemark[2] transformer model for comparative analysis. This method facilitates task-specific learning and offers a state-of-the-art benchmark for transformer-based sentiment analysis. The model utilizes dropout, dense layers, and ReLU activation to enhance regularization and feature transformation, making it an effective choice for sentiment classification tasks.

\footnotetext[2]{Available: \href{https://huggingface.co/nlptown/bert-base-multilingual-uncased-sentiment}{\texttt{https://huggingface.co/nlptown/bert-base-multilingual-uncased-\linebreak sentiment}}}

\subsection{Performance Evaluation}
The performance of each model is evaluated using several metrics to provide a comprehensive assessment. These metrics include accuracy, precision, recall, and F1 score. Accuracy measures the proportion of correctly classified instances. Precision assesses the accuracy of positive predictions, while recall measures the ability to identify all relevant instances. The F1 score, the harmonic mean of precision and recall, provides a balanced evaluation metric.

By comparing the models across different vectorization techniques, the best-performing model for each technique is determined and the overall best approach is selected based on a thorough performance analysis. This systematic comparison ensures the identification of the most effective model for emotion classification in Spanish customer feedback.

\section{Experiments}
\label{sec:experiments}

\subsection{Datasets}
The dataset we use for our experiments is sourced from customer feedback in the healthcare sector provided by the GreHus company. This dataset comprises responses to open-ended questions aimed at capturing a wide range of customer sentiments, including motivations, expectations, and suggestions for improvement. It contains 1045 entries, providing a comprehensive view of customer feedback in Spanish.

In addition to the original Spanish dataset, we create an English-translated version from the same Spanish dataset to evaluate model performance and assess the impact of translation on accuracy. The translation is performed using a MarianMT model, specifically the Helsinki-NLP/opus-mt-es-en model\footnotemark[3] from Hugging Face, which is designed for Spanish-to-English translation. The translation process involves tokenizing the text using the MarianTokenizer, generating translations with the MarianMTModel, and decoding the output to produce the English text. It contains 1045 entries.

\footnotetext[3]{Available: \href{https://huggingface.co/Helsinki-NLP/opus-mt-es-en}{https://huggingface.co/Helsinki-NLP/opus-mt-es-en}}

\subsection{Experiment Setup}
The experimental setup is designed to systematically evaluate the performance of various ML models in classifying emotions within the customer feedback data, as well as assess the effectiveness of our ensemble approach, CSE. 

\subsubsection{Dataset} The data is preprocessed by dropping duplicates, converting text to lowercase, and checking missing values to standardize and clean it for analysis. To transform the textual data into numerical representation, we use TF-IDF and TF-IDF with BERT embeddings. We utilize a pre-trained BERT model with padding and truncation applied to ensure a consistent sequence length of 512 tokens.

Combining these techniques helps capture the importance of words and deeper semantic contexts. TF-IDF captures statistical information, whereas BERT captures contextual semantic information. Combining TF-IDF and BERT vectors aims to enhance the models' ability to understand and classify emotions accurately. Subsequently, the dataset is split into training and testing sets with an 80/20 ratio.

\subsubsection{Model} Logistic Regression is a widely use baseline model for text classification. To ensure convergence, the maximum number of iterations is set to the best value obtained from the hyperparameter grid search. Logistic Regression is a linear model known for its interpretability and efficiency in binary and multi-class classification tasks. All other parameters are default.

A Bagging classifier is employed with LGBM as the base estimator. LGBM, a highly efficient gradient boosting framework, is chosen due to its ability to handle large-scale datasets and its inherent support for text classification tasks by dealing with high-dimensional data. The Bagging ensemble is created base learners (n\_estimators) obtained from grid search to ensure a balance between model diversity and computational efficiency.

AdaBoost is used as an ensemble technique to enhance model performance by sequentially improving the weak learners. For this task, decision trees are selected as the base learners, with the number of boosting stages set based on value from grid search (n\_estimators). AdaBoost adapts to the misclassified instances by assigning higher weights to these instances in subsequent rounds, thus refining the classifier with each iteration. This technique is particularly effective for text data, where distinguishing between neutral and borderline sentiments can be challenging.

The KNN algorithm is applied with neighbors~(n\_neighbors) obtained from the grid search. KNN is a non-parametric, instance-based learning algorithm where predictions are based on the proximity of a data point to its nearest neighbors in feature space, calculated using the Euclidean distance. The choice of k is made to strike a balance between bias and variance.

The ensemble, CSE, uses base models, including Logistic Regression, Bagging classifier with LGBM, KNN, and AdaBoost, each chosen for its complementary strengths in text classification. These models are trained to handle the multi-class sentiment classification task. A one-vs-rest Logistic Regression model is selected as the meta-model to combine the predictions from the base models. The stacking classifier is trained using the TF-IDF and full feature set, which combines TF-IDF and BERT embeddings, ensuring the model captures both statistical and contextual information. 5-fold cross-validation is employed during training to ensure robustness, and the final model's performance is evaluated using accuracy and classification metrics on the test set.

\subsection{Results}

\subsubsection{Spanish Dataset}
We report the classification performance of several models on the Spanish dataset, using both TF-IDF and TF-IDF with BERT embeddings as feature-generation techniques, as shown in Table~\ref{tab:spanish_dataset}. The models are evaluated on key performance metrics, including accuracy, precision, recall, and F1-score.

\begin{table}[ht]
\centering
\caption{Model Performance Comparison for TF-IDF and TF-IDF + BERT with Spanish Dataset} \label{tab:spanish_dataset}
\begin{tabular}{|>{\centering\arraybackslash}m{4cm}|>{\centering\arraybackslash}m{2cm}|>{\centering\arraybackslash}m{2cm}|>{\centering\arraybackslash}m{2cm}|>{\centering\arraybackslash}m{2cm}|}
\hline
\textbf{Model} & \textbf{Accuracy} & \textbf{Precision} & \textbf{Recall} & \textbf{F1-Score} \\
\hline
\multicolumn{5}{|c|}{\textbf{TF-IDF}} \\
\hline
Logistic Regression & 0.85 & 0.86 & 0.85 & 0.85 \\
\hline
Bagging Classifier with LGBM & 0.86 & 0.87 & 0.86 & 0.86 \\
\hline
KNN & 0.70 & 0.70 & 0.71 & 0.70 \\
\hline
AdaBoost & 0.70 & 0.77 & 0.70 & 0.71 \\
\hline
CSE & \textbf{0.88} & 0.88 & 0.87 & 0.87 \\
\hline
\multicolumn{5}{|c|}{\textbf{TF-IDF + BERT}} \\
\hline
Logistic Regression & 0.88 & 0.88 & 0.88 & 0.88 \\
\hline
Bagging classifier with LGBM & 0.87 & 0.87 & 0.87 & 0.87 \\
\hline
KNN & 0.83 & 0.84 & 0.83 & 0.83 \\
\hline
AdaBoost & 0.80 & 0.83 & 0.80 & 0.81 \\
\hline
CSE & \textbf{0.93} & 0.93 & 0.93 & 0.93 \\
\hline
\end{tabular}
\end{table}

The classification models using TF-IDF as the feature extraction method show varying levels of performance. Among them, the CSE model outperforms the others, achieving an accuracy of~0.88, precision of~0.88, recall of~0.87, and F1-score of~0.87.

Logistic Regression performs similarly with an accuracy of~0.85, while the Bagging classifier with LGBM and KNN exhibits moderate performance. AdaBoost and KNN perform the worst using only TF-IDF features, with accuracies around~0.70.

However, when combining TF-IDF with BERT embeddings, all models show improved performance. CSE stands out with an accuracy of~0.93, followed by Logistic Regression with~0.88 and Bagging classifier with~0.87 accuracy. The incorporation of BERT embeddings significantly enhances the contextual understanding of the models, leading to better classification results.

To further validate model performance, we conduct cross-validation using both Stratified K-fold and K-fold techniques for the training dataset as shown in Table~\ref{tab:cross_validation_es}. The results for both methods are quite similar, indicating the models are stable across validation strategies.

\begin{table}[ht]
\centering
\caption{Cross-Validation and Feature Generation Performance with Spanish Dataset} \label{tab:cross_validation_es}
\resizebox{\textwidth}{!}{ 
\begin{tabular}{|>{\centering\arraybackslash}m{4cm}|>{\centering\arraybackslash}m{2cm}|>{\centering\arraybackslash}m{2cm}|>{\centering\arraybackslash}m{2cm}|>{\centering\arraybackslash}m{2cm}|}
\hline
\textbf{Cross-Validation} & \textbf{Accuracy} & \textbf{Precision} & \textbf{Recall} & \textbf{F1-Score} \\
\hline
Stratified K-fold & 0.89 & 0.89 & 0.89 & 0.89 \\
\hline
K-fold & 0.88  & 0.89  & 0.88  & 0.88  \\
\hline
\end{tabular}}
\end{table}

The Stratified K-fold technique yielded an accuracy of~0.89, precision of~0.89, recall of~0.89, and F1-score of~0.89, while the K-fold method produced an accuracy of~0.88, precision of~0.89, recall of~0.88, and F1-score of~0.88. These consistent results across different cross-validation methods indicate the robustness and generalizability of the models on the Spanish dataset.


\subsubsection{English Translated Dataset}
We present the performance results of various models applied to the English-translated dataset, using both TF-IDF and TF-IDF with BERT embeddings with results displayed in Table~\ref{tab:english_translated}. The evaluation metrics include accuracy, precision, recall, and F1-score.

\begin{table}[ht]
\centering
\caption{Model Performance Comparison for TF-IDF and TF-IDF + BERT with English Translated Dataset} \label{tab:english_translated}
\begin{tabular}{|>{\centering\arraybackslash}m{4cm}|>{\centering\arraybackslash}m{2cm}|>{\centering\arraybackslash}m{2cm}|>{\centering\arraybackslash}m{2cm}|>{\centering\arraybackslash}m{2cm}|}
\hline
\textbf{Model} & \textbf{Accuracy} & \textbf{Precision} & \textbf{Recall} & \textbf{F1-Score} \\
\hline
\multicolumn{5}{|c|}{\textbf{TF-IDF}} \\
\hline
Logistic Regression & 0.85 & 0.86 & 0.85 & 0.85 \\
\hline
Bagging classifier with LGBM & 0.77 & 0.80 & 0.78 & 0.78 \\
\hline
KNN & 0.75 & 0.76 & 0.76 & 0.75 \\
\hline
AdaBoost & 0.71 & 0.75 & 0.71 & 0.72 \\
\hline
CSE & 0.87 & 0.88 & 0.87 & 0.87 \\
\hline
\multicolumn{5}{|c|}{\textbf{TF-IDF + BERT}} \\
\hline
Logistic Regression & 0.87 & 0.88 & 0.87 & 0.87 \\
\hline
Bagging classifier with LGBM & 0.86 & 0.86 & 0.86 & 0.86 \\
\hline
KNN & 0.85 & 0.85 & 0.85 & 0.85 \\
\hline
AdaBoost & 0.80 & 0.81 & 0.80 & 0.80 \\
\hline
CSE & \textbf{0.88} & 0.88 & 0.88 & 0.88 \\
\hline
\end{tabular}
\end{table}

The results obtained using TF-IDF as the feature extraction technique show that the CSE model delivers the best performance with an accuracy of~0.87, precision of~0.88, recall of~0.87, and an F1-score of~0.87.

Logistic Regression performs relatively well with an accuracy of~0.85, while the Bagging classifier performs moderately well with an accuracy of~0.77. Both KNN and AdaBoost exhibit lower performance, with accuracies of~0.75 and~0.71, respectively.

However, when combining TF-IDF with BERT embeddings, all models improve their performance significantly. The CSE model achieves the highest performance with an accuracy of~0.88, followed closely by Logistic Regression at~0.87. The Bagging classifier also improves to an accuracy of~0.86, and both KNN and AdaBoost see improvements with accuracies of~0.85 and~0.80, respectively.

To further validate the models' performance, we conduct cross-validation using both Stratified K-fold and K-fold techniques. The results, shown in Table~\ref{tab:cross_validation_en}, demonstrate consistent performance across validation techniques, confirming the robustness of the models.

\begin{table}[ht]
\centering
\caption{Cross-Validation Performance Comparison with English Translated Dataset} \label{tab:cross_validation_en}
\begin{tabular}{|>{\centering\arraybackslash}m{4cm}|>{\centering\arraybackslash}m{2cm}|>{\centering\arraybackslash}m{2cm}|>{\centering\arraybackslash}m{2cm}|>{\centering\arraybackslash}m{2cm}|}
\hline
\textbf{Cross-Validation} & \textbf{Accuracy} & \textbf{Precision} & \textbf{Recall} & \textbf{F1-Score} \\
\hline
Stratified K-fold & 0.89  & 0.89  & 0.89  &  0.89  \\
\hline
K-fold & 0.89  & 0.90  & 0.89  & 0.89  \\
\hline
\end{tabular}
\end{table}

The Stratified K-fold technique results in an accuracy of~0.89, with precision, recall, and F1-scores all at~0.89, whereas the K-fold method achieved similar performance with a slightly higher precision of~0.90. These results demonstrate the stability and robustness of the models when applied to the English-translated dataset.

\subsubsection{Comparison} 
\begin{figure}[h]
    \centering
    \includegraphics[width=0.7\linewidth]{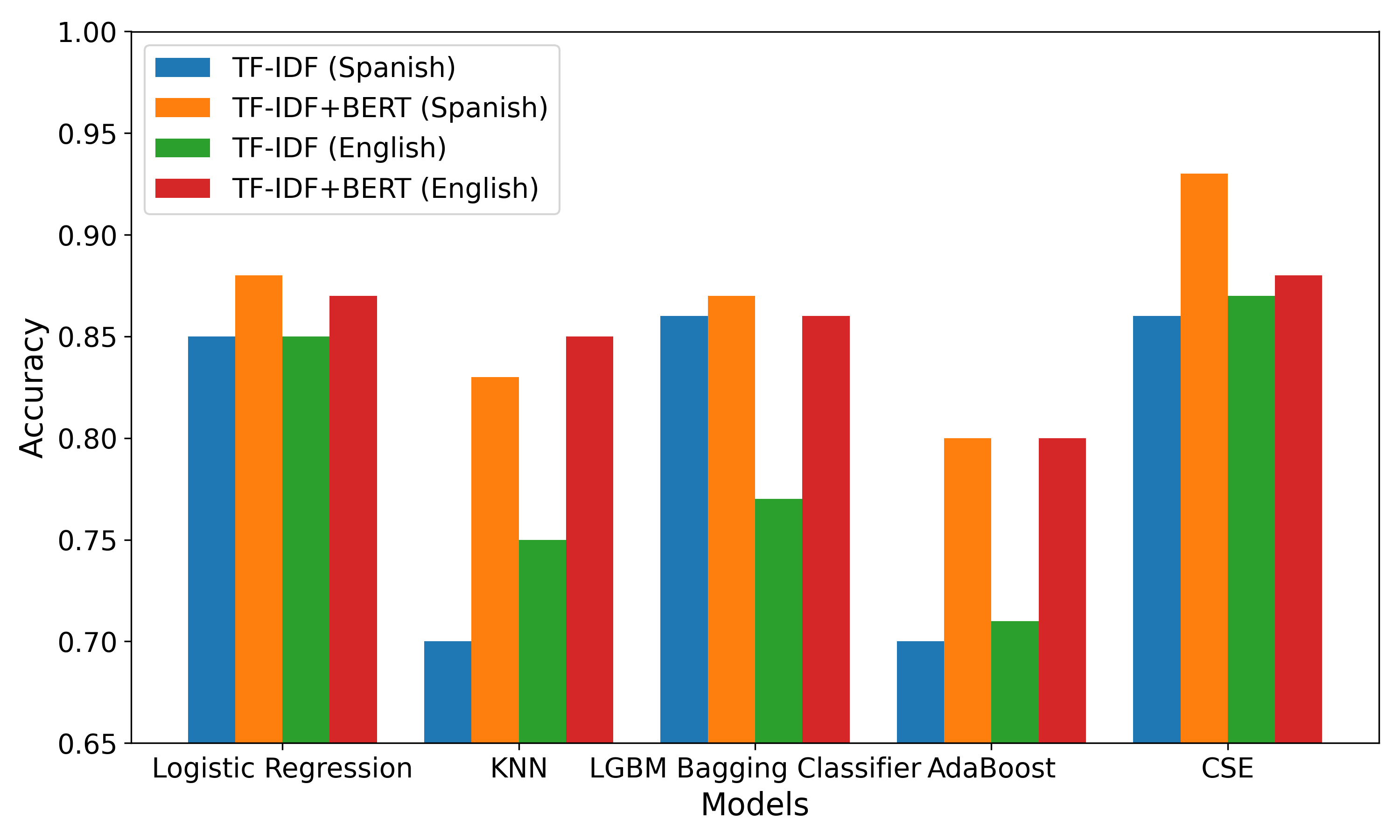}
    \caption{Model Accuracy Comparison for Spanish and English-translated Datasets}
    \label{fig:comparison}
\end{figure}

The comparison of the model performance between the Spanish and English-translated datasets, as illustrated in Figure~\ref{fig:comparison}, highlights significant differences in accuracy across models. The CSE consistently outperforms the other models in both datasets, with the Spanish dataset achieving the highest accuracy of~0.93 when using the combined TF-IDF with the BERT feature extraction technique. In comparison, the English-translated dataset reaches an accuracy of~0.88 using the same technique. Notably, all models show marked improvements in performance when TF-IDF is combined with BERT, with the Spanish dataset generally yielding higher accuracies than the English-translated one. This highlights that the models perform more accurately on the native Spanish dataset than the translation.

\begin{figure}[h]
    \centering
    \begin{subfigure}{0.49\textwidth}
        \centering
        \includegraphics[width=\linewidth]{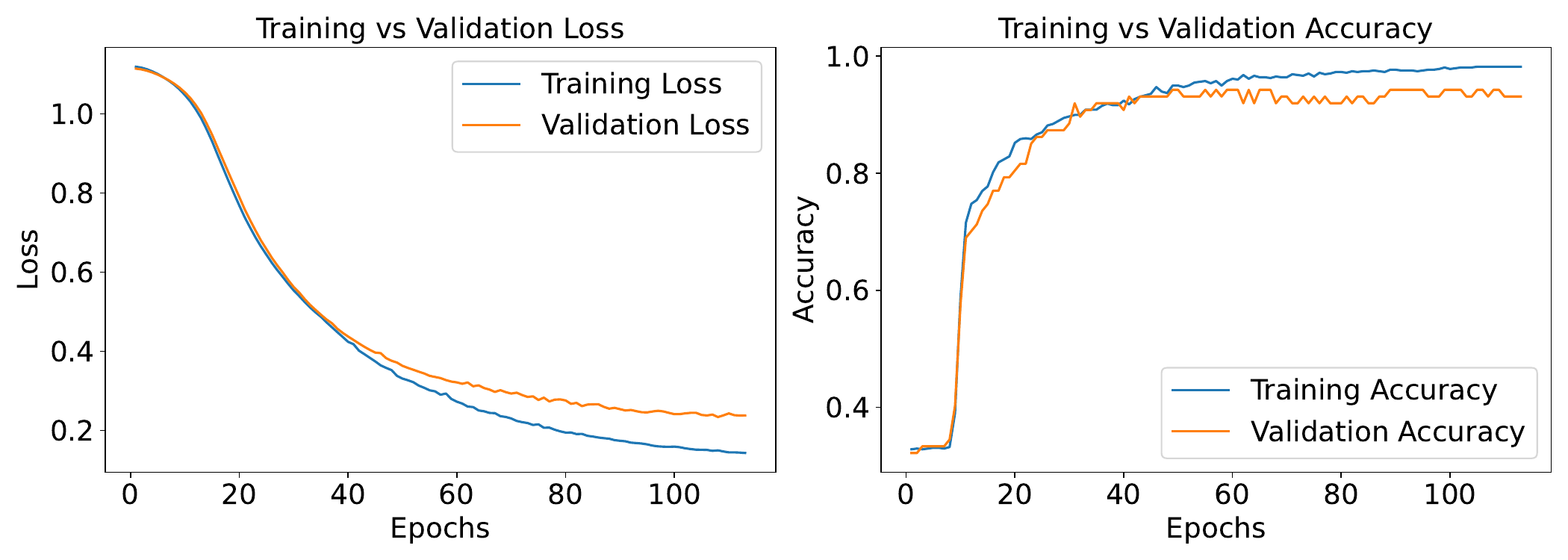}
        \caption{Spanish dataset history plot}
        \label{fig:spanish_history}
    \end{subfigure}
    \hfill
    \begin{subfigure}{0.49\textwidth}
        \centering
        \includegraphics[width=\linewidth]{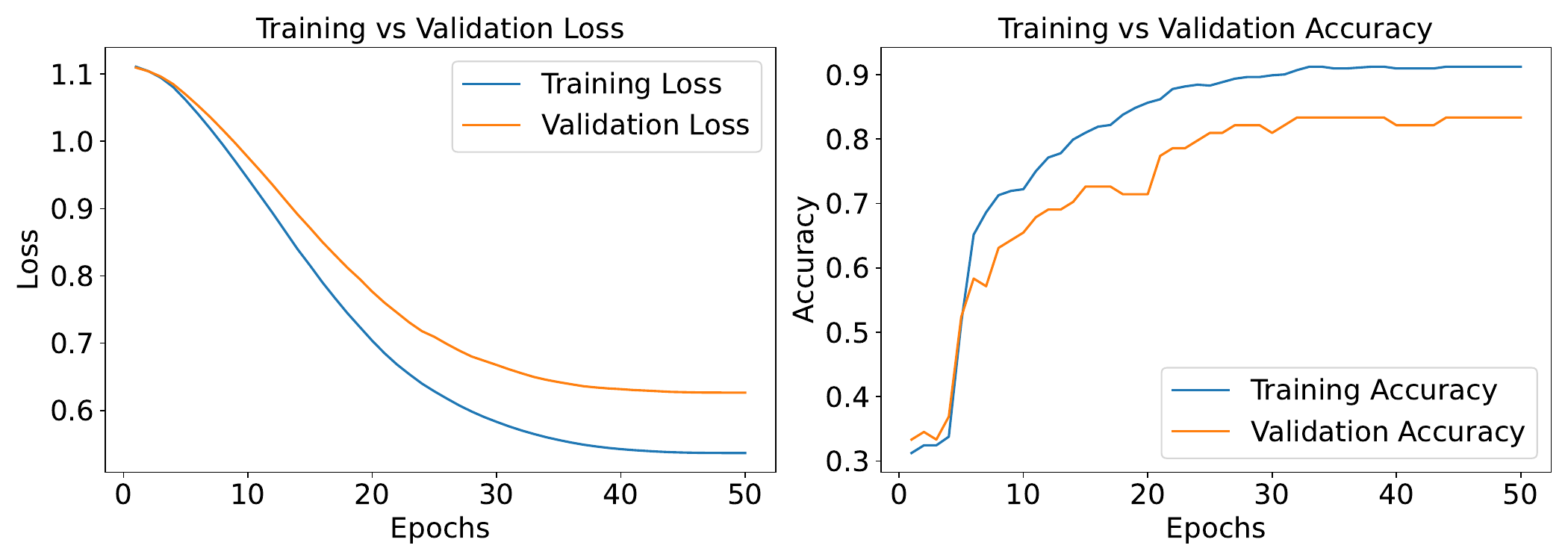}
        \caption{English dataset history plot}
        \label{fig:english_history}
    \end{subfigure}
    \caption{History plots for BERT model on Spanish and English datasets.}
    \label{fig:history_plot}
\end{figure}

The performance results of the state-of-the-art nlptown BERT sentiment analysis transformer model are shown in Table~\ref{tab:history_plot_comparison}. The model achieves a test accuracy of 0.89, along with precision, recall, and F1 scores of 0.9 each for the Spanish dataset. Similarly, we achieve a test accuracy of 0.86 for the English dataset, with precision, recall, and F1 scores of 0.86 each. Figure~\ref{fig:history_plot} presents the history plots for both the Spanish and English datasets using the nlptown BERT sentiment analysis model.

\begin{table}[ht]
\centering
\caption{Performance Comparison of CSE Approach and nlptown BERT Model} \label{tab:history_plot_comparison}
\begin{tabular}{|>{\centering\arraybackslash}m{4cm}|>{\centering\arraybackslash}m{2cm}|>{\centering\arraybackslash}m{2cm}|>{\centering\arraybackslash}m{2cm}|>{\centering\arraybackslash}m{2cm}|}
\hline
\textbf{Dataset} & \textbf{Accuracy} & \textbf{Precision} & \textbf{Recall} & \textbf{F1-Score} \\
\hline
Spanish Dataset & 0.89  & 0.90  & 0.90  & 0.90  \\
\hline
English Dataset & 0.86  & 0.86  & 0.86  & 0.86  \\
\hline
\end{tabular}
\end{table}



The comparison between the CSE ensemble approach and the BERT model highlights the impact of dataset size and task complexity. With a limited dataset, the data is insufficient for BERT to fully utilize its capabilities, as it typically requires significantly larger datasets to perform effectively. For simpler tasks like sentiment analysis, our approach, CSE, proves to be a better choice with higher accuracy, as it is simpler, more efficient, less resource-intensive, and less prone to overfitting. In contrast, BERT’s complexity makes it unnecessarily powerful and impractical for such tasks, as it is both inefficient and resource-intensive.

\section{Conclusion}
\label{sec:conclusion}

In this paper, we classify Spanish customer feedback into positive, neutral, and negative emotions using several machine learning models, with a focus on a CSE method. Our results show that the CSE, leveraging TF-IDF with BERT embeddings, significantly outperforms individual models in terms of accuracy, particularly when trained on native Spanish data. This underscores the importance of using original datasets to preserve linguistic nuances and achieve better classification accuracy. These results provide meaningful insights for enhancing customer sentiment analysis, and future research involves advancing models, such as transformer-based architecture, incorporating more comparison along with expanding datasets.


\section*{Acknowledgement}
We thank Grehus SAS BIC for providing anonymized data and express our gratitude to Eng. Javier Rojas, CEO, for his valuable insights and discussions. We also acknowledge Prof. Xi Peng for his guidance and constructive feedback.

\end{document}